 \documentclass[final,3p,times,twocolumn]{elsarticle}

\usepackage{graphicx}
\usepackage{amssymb}

\usepackage{lineno}
\usepackage{color}
\newcommand{\fvc}[1]{\textcolor{black}{#1}}
\usepackage{hyperref}

\journal{Journal of Neuroengineering and Rehabilitation}

\begin{document}

\begin{frontmatter}


\title{On Neuromechanical Approaches for the Study of Biological and Robotic Grasp and Manipulation}



\author{Francisco J Valero-Cuevas$^{1,2,*}$ and Marco Santello$^3$}

\address{1 Biomedical Engineering Department and \\
2 Division of Biokinesiology \& Physical Therapy\\
University of Southern California, Los Angeles, CA, USA\\
3 School of Biological and Health Systems Engineering\\ Arizona State University, Tempe, AZ, USA\\
$^*$ corresponding author
}

\begin{abstract}
Biological and robotic grasp and manipulation are undeniably similar at the level of mechanical task performance. However, their underlying fundamental biological vs. engineering mechanisms are, by definition, dramatically different and can even be antithetical. Even our approach to each is diametrically opposite: inductive science for the study of biological systems vs. engineering synthesis for the design and construction of robotic systems. The past 20 years have seen several conceptual advances in both fields and the quest to unify them. Chief among them is the reluctant recognition that their underlying fundamental mechanisms may actually share limited common ground, while exhibiting many fundamental differences. This recognition is particularly liberating because it allows us to resolve and move beyond multiple paradoxes and contradictions that arose from the initial reasonable assumption of a large common ground. Here, we begin by introducing the perspective of neuromechanics, which emphasizes that real-world behavior emerges from the intimate interactions among the physical structure of the system, the mechanical requirements of a task, the feasible neural control actions to produce it, and the ability of the neuromuscular system to adapt through interactions with the environment. This allows us to articulate a succinct overview of a few salient conceptual paradoxes and contradictions regarding under-determined vs. over-determined mechanics, under- vs. over-actuated control, prescribed vs. emergent function, learning vs. implementation vs. adaptation, prescriptive vs. descriptive synergies, and optimal vs. habitual performance. We conclude by presenting open questions and suggesting directions for future research. We hope this frank and open-minded assessment of the state-of-the-art will encourage and guide these communities to continue to interact and make progress in these important areas at the interface of neuromechanics, neuroscience, rehabilitation and robotics.
\end{abstract}


\end{frontmatter}


\section*{Introduction}
Grasp and manipulation have captivated the
imagination and interest of thinkers of all stripes over the millennia; and with enough
reverence to even attribute the intellectual evolution of humans to the capabilities
of this appendage \cite{Nussbaum1985Aristotle,wilson2010hand,bell1865hand}. Simply put, manipulation function is one of the key elements of our identity as a species (for an
overview see \cite{Valero-Cuevas2009Why}). This is a natural response to the
fact that much of our physical and cognitive ability and well-being is intimately
tied to the use of our hands. Importantly, we have shaped our tools and
environment to match its capabilities (straightforward examples include lever
handles, frets in string instruments, and touch-screens). This co-evolution between
hand-and-world reinforces the notion that our hands are truly amazing and robust
manipulators, as well as providing rich sensory, perceptual and even social information.
 
It then comes as no surprise that engineers and physicians have long sought to replicate and restore this functionality in machines---both as appendages to robots and prostheses attached to humans with missing upper limbs \cite{cordella2016literature}. Robotic hands and prostheses have a long and illustrious history\fvc{, with records of sophisticated articulated hands as early as Gottfried `G{\"o}tz' von Berlichingen's iron hand in 1504 \cite{loffler1984ersatz}. Other efforts \cite{thurston2007pare,norton2007brief,hillman20042,schlesinger1919mechanische,kluge1842arthroplastik} were often} fueled by the injuries of war \cite{panchasi1995reconstructions,serlin2002artificial,mcsorley2012war,ling2010surgical} and the Industrial Revolution \cite{riskin2007eighteenth}. The higher survival rate in soldiers who lose upper limbs \cite{feinglass2001postoperative, kristensen2012very} and the continual emergence of artificial intelligence \cite{mccorduck1977history,benko2009history} are but the latest impetus. Thus, the past 20 years have seen an explosion in designs, fueled by large scale governmental funding (e.g., DARPA's Revolutionizing Prosthetics and HAPTIX projects, EU's INPUT and SOFTPRO projects) and private efforts such as DeepMind. A new player in this space are the potentially more revolutionary social networks of high-quality amateur scientists as exemplified by the FABLAB movement \cite{gershenfeld2008fab}. They are enabled by ubiquitously accessible and inexpensive 3D printing and additive manufacturing tools \cite{prince20143d}, collaborative design databases (\url{www.eng.yale.edu/grablab/openhand/} and others), and communities with formal journals (\url{www.liebertpub.com/overview/3d-printing-and-additive-manufacturing/621/} and \url{www.journals.elsevier.com/additive-manufacturing/}). Grassroots communities have also emerged that can, for example, compare and contrast the functionality of prosthetic hands whose price differs by three orders of magnitude (\url{3dprint.com/2438/50-prosthetic-3d-printed-hand/}).
 
For all the progress that we have seen, however, (i) robotic platforms remain best at pre-sorted, pick-and-place assembly tasks \cite{Honarpardaz2017104}; and (ii) many prosthetic users still prefer simple designs like the revered whole- or split-hook designs originally developed centuries ago \cite{weir2003design,childress1985historical}.

Why have robotic and prosthetic hands not come of age? This short review provides a current attempt to tackle this long-standing question in response to the current technological boom in robotic and prosthetic limbs. Similar booms occurred in response to upper limb injuries \cite{childress1985historical} after the Napoleonic \cite{wellerson1958manual}, First \cite{panchasi1995reconstructions}, and Second World Wars \cite{norton2007brief}, and---with the advent of powerful inexpensive computers---in response to industrial and space exploration needs in the 1960's, 1970's and 1980's \cite{cutkosky1989grasp,colgate1988robust,murray1994mathematical,uchiyama1990SpaceArm,aikenhead1983canadarm,nguyen1989NASAdyn}. We argue that a truly bio-inspired approach suffers, by definition, from both gaps in our understanding of the biology, and technical challenges in mimicking (what we understand of) biological sensors, motors and controllers. Although biomimicry is often not the ultimate goal in robotics in general, it is relevant for (1) humanoids and (2) prostheses. Thus, our approach is to clarify when and why a better understanding of the biology of grasp and manipulation would benefit robotic grasping and manipulation.

Similarly, why is our understanding of the nature, function and rehabilitation of biological arms and hands incomplete? Jacob Benignus Winsl\o w (Jacques-B\'enigne Winslow, 1669---1760) noted in his \emph{Exposition anatomique de la structure du corps humain} (1732) that `The coordination of the muscles of the live hand will never be understood' \cite{winslow1732exposition}. Interestingly, he is still mostly correct. As commented in detail before \cite{Valero-Cuevas2009Why}, there has been much work devoted to inferring the anatomical, physiological, neural and cognitive processes that produce the upper limb function we so dearly appreciate and passionately work to restore following trauma or pathology. We argue, as Galileo Galilei did, that mathematics and engineering have much to contribute to the understanding of biological system. Without such a `mathematical language,' we run the risk, as Galileo put it, of `wandering in vain through a dark labyrinth' \cite{galilei1623saggiatore}. Thus, this short review also attempts to point out important mathematical and engineering developments and advances that have helped our understanding of our hands.

This review first contrasts the fundamental differences between engineering and neuroscience approaches to biological vs. robotic systems. Whereas the former applies engineering principles, the latter relies on scientific inference. We then discuss how the physics of the world provides a common ground between them because both types of systems have similar functional goals, and must abide by the same physical laws. We go on to evaluate how biological and robotic systems implement the necessary sensory and motor functions using the dramatically different anatomy, morphology and mechanisms available to each. This inevitably raises questions about differences in their sensorimotor control strategies. Whereas engineering system can be designed and manufactured to optimize well-defined functional features, biological systems evolve without such strict tautology. Biological systems likely evolve by implementing ecologically and temporally good-enough, sub-optimal or habitual control strategies in response to the current multi-dimensional functional constraints and goals in the presence of competition, variability, uncertainty, and noise. We conclude by exploring the notion that the functional versatility of biological systems that roboticists admire is, in fact, enabled by the very nonlinearities and complexities in anatomy, sensorimotor physiology, and neural function that engineering approaches often seek to avoid.

\section*{\fvc{Deductive and inductive} science for the study of biological systems vs. engineering synthesis for the construction of robotic systems}
 
As we have discussed before \cite{valero-cuevas2009computational}, any understanding of a biological system is done by a difficult scientific process of \fvc{logical} inference \cite{krakauer2017neuroscience} \footnote{\fvc{Quoted from \cite{WikipediaDeductiveReasonings2017}: `Deductive reasoning (top-down logic) contrasts with inductive reasoning (bottom-up logic) in the following way: In deductive reasoning, a conclusion is reached reductively by applying general rules that hold over the entirety of a closed domain of discourse, narrowing the range under consideration until only the conclusion(s) is left.'. Conversely \cite{WikipediaInductiveReasonings2017}, 'Inductive reasoning is reasoning in which the premises are viewed as supplying strong evidence for the truth of the conclusion. While the conclusion of a deductive argument is certain, the truth of the conclusion of an inductive argument may be probable, based upon the evidence given \cite{copi2016essentials}.'} }.  \fvc{Scientific inquiry, in particular, is a combination of a deductive (top-down logic) approach that invokes laws of physics---with inductive (bottom-up logic) reasoning that uses specific instances of} observed behavior in the complete system (e.g., gait, flight, manipulation) to build conceptual, analytical or computational models (i.e., hypotheses) about how constitutive parts interact to produce the overall behavior. In contrast, the engineering perspective is to use proven physical laws (i.e., mechanisms) to synthesize and build complex systems, like robots, in a bottom-up way. The multiple technological successes of this engineering approach naturally encourage us to identify such fundamental mechanisms in biological systems, and assemble models and hypotheses about how they interact in biological systems to explain vertebrate function and dysfunction---and apply them to revolutionize rehabilitation medicine and build better machines that are either bio-inspired or able to interact with humans (e.g., exoskeletons).

Thus, this engineering bottom-up approach has been used by biomechanists, kinesiologists, clinicians, and computational- and neuro-scientists to build models to address questions such as `What is the control strategy the nervous system uses to (i) move the arm to place the hand in a sequence of locations in space? Or (ii) reach to an object and grasp an object? Or (iii) use the fingers to use a hand held tool?'  Similarly, 'How do specific parts of the brain contribute to produce the specific features of reach, grasp and manipulation?' (for recent reviews see, e.g.,  \cite{Gandevia2011constraints, Valero-Cuevas2009Why, santello2015getting, jacobs2010human, davare2011interactions,lemon2008descending}).

\fvc{For example, neuroscientists have characterized the cortical networks responsible for selection of hand postures, forces or dexterous manipulation \cite{mosier2011controlling,ejaz2015hand}. Specifically, invasive and noninvasive recordings from cortical regions in non-human and human primates, and non-invasive brain stimulation studies in humans have revealed the functional role of sensory, premotor, motor, and associated areas in motor planning, sensorimotor adaptation, grip type selection, storing and retrieving sensorimotor memories of hand-object interactions, and controlling grasping and manipulation (for review see\cite{ davare2011interactions}).
}

However, inferring valid models of biological systems is not trivial. It remains reasonable to ask whether it is even possible for \fvc{us} to produce robust insights about the mechanisms underlying complex neuromuscular systems \cite{jonas2017microprocessor}. \fvc{Scientific inquiry} requires that we trust current (imperfect) theories of the mechanisms behind the material properties, sensors, muscles, and neural processing in biological systems which, when interacting \fvc{with physical laws} in a particular functional regime (e.g., turbulent and laminar flow, continuum and rigid body mechanics, stable and unstable dynamical domains, information theory, etc.), give rise to the observed biological behavior. Unfortunately, the differences that invariably emerge between model predictions and experimental data can be attributed to a variety of sources, ranging from the validity of the scientific hypothesis being tested to the choice of each constitutive element, or even their numerical implementation \cite{Valero-Cuevas2007Beyond,valero-cuevas2009computational}. Even when conceptual, analytical or computational models are carefully assembled, the modeler must make arbitrary choices that often affect the predictions of the model in counterintuitive ways. Some examples of unavoidable choices are the types of models for joints (e.g., a hinge vs. articulating surfaces), muscles (e.g., simple low pass filters, Hill-type, populations of motor units), controllers (e.g., proportional-derivative, Bayesian, internal models, optimal control), and solution methods (e.g., forward, inverse) \cite{valero-cuevas2009computational}. These choices are driven, at best, by a comprehensive distillation of the vast literature, a focused research scope on simplified or specific scales or domains of function, or intentional sidestepping of unknown aspects of model elements and their interactions that cannot be easily or accurately measured experimentally---and thus, cannot be confidently included in the model. At worst, (over)simplifications are driven by biological/mathematical/computational convenience or expediency. 

\fvc{A salient example of applying scientific findings to robotic systems is the application of kinematic (postural) hand synergies to the design of the controller for a robotic hand, such as the Pisa/IIT SoftHand \cite{catalano2014adaptive}. This design is based on using the first principal component computed from a set of static hand postures recorded in human subjects while grasping imagined objects \cite{santello1998postural}. Here, the observed covariation in joint angles of the digits was captured by the first principal components accounting for over 50\% of variance of kinematic data. This inspired the robotic design of the Pisa/IIT SoftHand where a single actuator drives the motion across multiple digits along that main joint-angle coordination pattern.}

\fvc{It is important to note that it is perhaps the compliance built into this robotic hand by design that allows a passive (i.e., uncontrolled) adaptation to the specifics of each grasp \cite{brock2016transferring}. Such counter-intuitive and often overlooked contributions of `passive' structures to static and dynamic functional versatility falls within the realm of morphological computation---a longstanding concept that is being revisited as an important contributor to the versatility of biological systems, robots and prosthetics (e.g., \cite{ pfeifer2009morphological,valero2007tendon, alexander1968animal,mardula2015implanted,odhner2014compliant}).}

\fvc{The \emph{status quo} in traditional robotics is a prime example of the converse, i.e., engineering synthesis. That is, building robotic systems that are the embodiment and application of the closed-form mathematical, design, and control principles we have and understand well. To give an example discussed in detail below, many robots are designed to have rotational motors. This comes from the fact that the torque-control formulation for robotics is well developed \cite{valero2015fundamentals}.}

\fvc{However, the \emph{control} of robotic systems can also depend heavily on a deductive approach. Namely, most control architectures include an idealized mathematical model of the system they are controlling (i.e., the so-called plant\footnote{A note on nomenclature. This does not mean to tip over a potted houseplant. This term comes from control engineering where the process to be controlled was usually an industrial or manufacturing plant.}) \cite{ogata1997,stengel2012optimal}. It is common practice to, for example, use system identification techniques \cite{verhaegen2007filtering,ljung1999system} to use experimental data to infer a data-driven model, i.e. input/output transfer function of the plant `as built’ \cite{van2002identification,jalaleddini2017subspace,perreault1999multiple}. These models range from linear models through to nonlinear and probabilistic approximations to the robotic system’s dynamics \cite{hollerbach1995closed,bobrow1998modeling,johansson2000state,lewis1998neural,ortega1989adaptive}.
This is especially necessary when using optimal control formulations which depend critically on an accurate model of the plant \cite{doyle1978}. These limitations are currently being addressed by real-time model predictive control strategies that continually update families of possible models of the plant (and its state), and operate only over a limited time horizon, e.g.,  \cite{williams2016aggressive}.}

\section*{\fvc{Mechanics and neuromechanics as the common ground between biological and robotic systems for grasp and manipulation}}
The fields of biological and robotic behavior are, nevertheless, fortunate in that principles of mechanics are at the root of both evolutionary biology and robotics. Darwinian evolution and Newtonian mechanics are unforgiving arbiters that continually shape what is possible and successful in the physical world. Thus, even though animals have had to evolve whereas robots have had to be designed and built, both had to successfully withstand and exploit the laws of mechanics  \cite{valero2015fundamentals}. Therefore, studying biological systems in the context of the physical function of grasp and manipulation does have the possibility of providing insights into how the structure of the body, information processing in the nervous system, and their interactions give rise to complex behavior.

An appropriate name for this approach is \emph{Neuromechanics}. To our knowledge, this term was first coined by Enoka in his 1988 book \emph{Neuromechanical Basis of Kinesiology} \cite{enoka1988neuromechanical}. \fvc{We use the term neuromechanics as describing the functional co-adaptations of the nervous, motor, sensory and musculoskeletal systems to produce effective and versatile mechanical behavior with a physical body in the physical world.} It aptly emphasizes that, in a field that mostly emphasizes the cognitive capabilities of the mammalian brain, it is easy to overlook that the nervous system and body co-evolved well before mammals appeared \cite{lisberger2013principles}. Thus, there is much to be learned when studying neural function using physical behavior by the periphery (i.e., limbs) as a means to understand central (i.e., neural) function  \cite{valero2015fundamentals}. But how can we move away from the difficulties in deductive inference mentioned in the prior section?

\fvc{One promising approach is to use synthetic analysis to build neuromorphic neuromechanical systems that exploit physical reality as the common ground between biological and robotic systems. The neuromorphic approach reflects the sentiment expressed by  Richard Feynman}, `What I cannot build I do not understand. Know how to solve every problem that has been solved.' \footnote{This was found written on his blackboard at the time of his death in February 1988, \url{https://archives.caltech.edu/pictures/1.10-29.jpg}. \fvc{It is thought he meant to suggests that one should only use mathematical concepts one has derived, and therefore proven, to oneself.}} In our context, it can be taken to mean that, if we have over one hundred years of sensorimotor neuroscience since Sir Charles Sherrington \cite{sherrington1913reflex}, and if the principles we have deduced are sound, then we should be able to build components that embody those mechanisms in such a way that when assembled they behave like biological systems \cite{niu2017neuromorphic,jalaleddini2017neuromorphic}. \fvc{One example of such a neuromorphic approach} uses ultra-fast computer processors to simultaneously implement populations of autonomous, interconnected spiking neurons in real time that follow Hodgkin-Huxley rules of how action potentials in neurons are initiated and propagated \cite{hodgkin1952dual}. As mentioned in \cite{niu2017neuromorphic,olshausen1996emergence}, this general approach has \fvc{also} been successfully applied to understand mechanisms of memory, visual representation, and recently, cognitive function. Note that neuromorphic is distinct from neuromimetic or neuroinspired. Biomimetic (neuromimetic) and bioinspired (neuroinspired) work seeks to copy or replicate the biological (neural) behavior by any engineering means---like prosthetic hands that have no muscles or tendons, or airplanes that fly without flapping wings. In contrast, neuromorphic  approaches use engineering means to implement the biological mechanisms themselves.

As explained in \cite{niu2017neuromorphic},  we have taken this approach one step further by combining neuromorphic \emph{and} neuromechanical approaches, we seek to implement the the neural control of the body---effectively merging biology and robotics in the arena of physical function. We have coupled real-time neuromorphic implementations of stretch reflex circuitry in populations of spinal neurons, to electric motors controlled by real-time models of muscle function to apply forces to the tendons of actual human cadaveric fingers. This is the first neuromorphic neuromechanical system, to our knowledge, that has put our understanding of fundamental sensorimotor mechanisms in the spinal cord to the ultimate test of physical implementation. Importantly, the behavior emerges from the system as it is not prescribed beyond the nature and connectivity of its elements. An added advantage is that one can also `record' from single or multiple neurons, motor units, afferent nerves, etc. to explore emergent behavior at truly multiple scales. So far, this approach has allowed us to begin to understand cardinal features of afferent muscles of human fingers to replicate fundamental features of healthy muscle tone, hypo and hypertonia \cite{jalaleddini2017neuromorphic}.

While still an imperfect approximation, this neo-Sherringtonian approach helps us test arguments about which specific features of spiking neurons and their connectivity, spindle function, fusimotor drive, descending commands, finger anatomy and tendon/skin/joint tissue properties \emph{suffice} to produce realistic healthy and pathologic behavior in afferent muscles acting on anatomical fingers. Moreover, this has the advantage of  using physical behavior as the ground truth for the evaluation of functional performance.

\fvc{A combined neuromorphic and neuromechanical approach, although grounded and developed for neuroscientific applications, could potentially inspire robotics research and design by revealing insights into how complex behaviors emerge from adaptation of neural controllers to mechanical properties of physical systems.}
 
 \section*{Fundamental differences between biological and robotic systems}
\subsection*{Sensory differences}
From the perspective of biological vs. \fvc{robotic} closed-loop behavior, \fvc{a striking difference is the superior ability of the nervous system} to utilize and effectively integrate information acquired through an incredibly wide array of nonlinear, delayed, noisy, non-collocated and distributed sensors (for review, see \cite{johansson2009coding}). Important advances have been made in our understanding of how multimodal sensory information is integrated to make possible the ability of humans to extrapolate sensory information based on the statistical properties of stimuli \cite{ernst2002humans}, as well as its vulnerability to sensory illusions \cite{shibata2014digit}.

Remarkably, the hand's sensory system endows humans not only the ability to perform online sensing of the state of the system (e.g., contact onset and offset) \cite{wolpert2000computational,wolpert2001perspectives}, but also to use sensory information acquired through past hand-object interactions to predict sensory consequences stemming from planned interactions \cite{johansson1984roles,Johansson1988}. Even more impressive is the exquisite ability of vertebrates to perform `active sensing' (e.g., whisking in rodents \cite{celikel2007sensory,solomon2006biomechanics}) and tactile exploration in humans of the shape, texture, features and mechanical properties of objects \cite{lederman1987hand,lederman1993extracting} \footnote{Other important forms of active learning are active vision and saccades in mammalian vision \cite{olshausen201320years}}. In active sensing, motor actions are explicitly driven to extract relevant sensory information \cite{eggermann2014cholinergic}. This has led some to express that the human hand is as much a sensory organ as a motor effector \cite{vallbo1984properties}.

In robotics, sensory data are central to feedback control \cite{siciliano2010robotics}. The idea of extending the use of sensory data beyond feedback to also extract the properties of a system is called system identification \cite{ljung1999system} or `plant inversion' \cite{wolpert2000computational,uhl2007inverse}. That is, characterizing the effects of command signals (i.e., the input-output characteristics of the plant) can be done when the process is inverted mathematically (i.e., find the output-input characteristics), analytically or experimentally. Hence the term `plant inversion.'  System identification has been an extensive field since the middle of last century \cite{westwick2003identification,verhaegen2007filtering}, and continues to make progress as sensors, algorithms and processing power increase \cite{yang2016nonlinear,jalaleddini2013subspace,de2003closed,ludvig2012system}. 

An important challenge in tactile sensors is their problematic placement at the fingertips, where they are exposed to damage \cite{shamanna2006micromachined,lowe2004flexible}. A notable advance has been the biomimetic idea by Loeb and Johansson to develop the Biotac \cite{loeb2010biomimetic, wettels2008biomimetic}, where the fingertip itself is a rugged and ribbed rubber balloon (much like a finger pad) inflated with conductive fluid over the distal bone. Static and dynamic contacts with objects produce changes in hydrostatic pressure and electrical impedance in the fluid that produce a rich and time-varying multi-dimensional sensory signal. The commercial version of this system is now being incorporated into multiple robotic hands \cite{fishel2013syntouch}. It is even being deployed to create industrial standards for the perceptual aspects of touch such as smoothness, roughness, silkiness, richness, quality, etc. Solving the difficult signal processing challenges that this type of tactile sensor presents has the potential to imbue robotic hands with truly robust and useful synthetic touch. Solving the computational challenges of sensory fusion (let alone the active sensing) in robotic and prosthetic systems is a critical frontier. There are even efforts at the interface of prosthetics and robotics to translate touch information from prosthetic hands into neurostimulation to restore the sense of touch (e.g., \cite{horch2011object,tabot2013restoring,Raspopovic2014}).

\fvc{Artificial feedback approaches, of course, extend beyond these examples of sensors located on the fingertips, and include artificial skins with embedded strain gauges (e.g., \cite{kim2014stretchable}) and vibrotactile feedback used in prosthetics delivering information about mechanical events (i.e., contact) to the residual limb (e.g., \cite{antfolk2013artificial}).
}

\fvc{Nevertheless, the necessity of sensory information for manipulation has been challenged by practical examples of sensor-less, fully open-loop grasp \cite{Valero2007Anatomical,brock2016transferring,deimel2016novel,catalano2014adaptive,odhner2014compliant}, and pre-planned manipulation \cite{imai2004dynamiccathching}. It is therefore most likely that in biological systems---and by extension in robots---sensory information is most useful during learning \cite{wolpert2001perspectives} (see section  on learning below).}

\subsection*{Motor differences}

Chief among these is our inability, so far, to match the power:weight ratio, mechanical efficiency, versatility, adaptability and self-repair properties of muscle. A pneumatic analog to muscle was first developed by McKibben in the 1950's \cite{schulte1961characteristics,gavrilovic1969positional,chou1996measurement} and continues to be used and studied \cite{kodama2016simultaneous}\cite{gordon2006mechanical}, but the need for compressors/pressurized tanks, valves, cables, mufflers, etc. remains a challenge to its portability and versatility. Electric, hydraulic and pneumatic actuators are, of course, the mainstay of robotics. \fvc{The last two decades have seen great progress in the technology to control \cite{jackson2015experimental} and power useful and portable} exoskeletons and prostheses \cite{van2009compliant,pons2002high,bogue2009exoskeletons}. \fvc{Moreover, battery life has improved by several orders of magnitude} \cite{Rahman2014Battery,ScrosatiLiReview2010,YoshinoBookChapter}. However, muscles remain unmatched in the continually surprising variety of mechanical functions they accomplish in locomotion and manipulation; serving as motors, brakes, springs, struts, etc. \cite{dickinson2000animals,biewener2016locomotion}.

From the architectural perspective, contractile proteins in muscle can only make muscles actively pull on their tendons, which attach to bones after crossing joints \cite{lieber1992skeletal,enoka2008neuromechanics}. Consequently, limb function in vertebrates is tendon-driven, not torque-driven as in the mathematics and dominant practice of robotics.

A recent conceptual advancement in the study of sensorimotor \fvc{control} in vertebrates  comes from embracing \fvc{and emphasizing} the fact that muscles act on the body via tendons \cite{valero2015fundamentals}. While this has been obvious since the very first anatomical studies of antiquity, most modern engineering, neuroscience, biomechanical, and mathematical analyses have tended to prefer the torque-driven abstraction. To be fair, the torque-driven phrasing of the problem is attractive, mathematically correct, and well developed from the conceptual, analytical and computational perspectives. In it, rotational motors at each hinge joint produce torque, angle, or angular velocity directly, which give rise to the kinematics and kinetics of the limbs and fingers they control. The actions of muscles in a simulated biological system are, thus, collapsed into net joint torques at each joint. Then, their analysis proceeds as with any other robotic system. But there are several arguments concluding that this abstraction can be misleading; as it does not represent the actual mechanical problem of controlling tendons, which is the actual problem the nervous system confronts. Due to recent advances in the tendon-driven formulation of limb and finger function (e.g., \cite{martin2011constructing,valero2015fundamentals,Sueda2008,Kaufman2005,mao2012design,oh2005cable}), we are now better able to focus on the actual tendon-driven mechanical problem that confronts the nervous system. In fact, as argued in \cite{valero2015fundamentals}, many of the conceptual/mathematical problems associated with the analysis of the neuromuscular control of biological limbs can be clarified by using this perspective.

\subsection*{Sensorimotor \fvc{control}}
Due to the above-described limitations of \fvc{scientific} inference, it is not surprising that competing views exist on \emph{the} model (i.e., \emph{hypothesis, strategy}) that best explains the nervous system's exquisite ability to control movement, locomotion and manipulation; as well as its uncanny ability to generalize and adapt learned motor behaviors. \fvc{For example, the concepts of `internal models' and `optimal control' can capture significant features of motor behavior, but it has been challenging to associate these theoretical frameworks with specific anatomical structures or physiological mechanisms. } It is only natural that some of these models have been inspired by the formalism and successes of robotics and control theory, which in turn have been influential in driving experimental approaches and theoretical frameworks (for review, see \cite{Wolpert2011}). 

However, even a cursory comparison between how robotic and human hands reach, grasp and manipulate objects reveals major differences between them. Some differences stem from ability of the neuromuscular system to implement versatile transitions, adjustments and adaptations of control strategies. One example is the ability of the hand to swiftly change control strategies when transitioning from finger motion to force application \cite{Venkadesan2008Neural}. Another example is humans' ability to modulate finger force distribution shortly after contact and prior to onset of manipulation to account for trial-to-trial variability in finger placement \cite{Fu2010,Fu2011,Fu2014a,Fu2014,Marneweck2016}. Such problems are extremely challenging to replicate in a robotic system. Other differences emerge from the hand's unique anatomical structure, which allows it to adapt to task demands, including passively shaping itself to object geometry. These are differences that robotic designs can partially address (see section `Under- vs. Over-actuated control'), but cannot fully match given the limitations of robotic systems in actively integrating sensory feedback with motor commands, or passively adapting to objects and the environment.

\subsection*{\fvc{Advantages and limitations of control theoretic approaches to biological sensorimotor control}}

These fundamental differences motivate and justify a candid evaluation of the extent to which our conceptual approaches to robotics are appropriate for the study of sensorimotor \fvc{control} in biological systems.  As mentioned above, the approach to biological sensorimotor \fvc{control}  has, for historical and practical reasons, leveraged the formalism of robotics and control theory (e.g., \cite{murray1994mathematical,valero2015fundamentals,Yamaguchi1990,shadmehr2012biological}). At the risk of oversimplifying a large field for the sake of succinctness, we can describe real-time feedback control as follows: sensors transduce physical signals to estimate the performance of the system, which the controller considers as it applies control laws to take the next actions to correct mistakes, reject perturbations, and meet the constraints of the task to achieve a goal. It stands to reason that the neuromuscular control of biological systems can perform all of these functions.  Therefore, \fvc{control-theoretic} constructs likely apply (e.g., \cite{todorov2002optimal,scott2004optimal,loeb1999Hierarchical,loeb1990understanding,peterka2002sensorimotor}). Such an approach is justified by the successes in identifying neural circuits that perform closed-loop feedback control such as homeostasis in physiological control systems \cite{khoo2000physiological}, muscle stretch reflexes \cite{mcintyre1993servo}, and vestibulo–ocular reflexes for eye tracking in the presence of head rotation \cite{angelaki2008vestibular,ranjbaran2015hybrid}.

Why is it that such a well-founded approach has failed to produce conclusive theories for sensorimotor control in humans for grasp and manipulation?\footnote{One can argue that work in invertebrates or locomotor patterns in vertebrates has been more successful, e.g., \cite{suver2016sensory,harris2010general,grillner2006biological}.}  One possibility is that sensorimotor control for grasp and manipulation is unlike other forms of motor control because its motor actions involve perception, as well as more complex sensorimotor transformation processes than, say, cyclical movements for  locomotion. The importance of the interaction between perception and action has been convincingly argued by, for example, Prinz, Iberall and Arbib \cite{iberall1990schemes,mackenzie1994grasping,mechsner2001perceptual,charpentier1891analyse}. This is further exemplified by other work bridging this perception-action gap \cite{yue1992strength,gordon1993memory,murray1999charpentier,flanagan2000independence,warren2011effects,craje2013effects}.

This conceptual divergence across biological and robotics problems can perhaps be explained by assessing some fundamental features of the robotics perspective, and their appropriateness for the study of sensorimotor \fvc{control}  in biological systems. Earlier, we spoke of the \fvc{control-theoretic} framework that defines interactions among sensors, control laws and the goal of the task.  A more formal presentation is that the control laws operate to change the `state' of the system. The formal definition of the state of a system is the minimal set of variables to characterize its governing equations, or that can be used to describe its dynamical evolution \cite{ogata1997,bryson1975applied,murray1994mathematical}. Thus, control theory is based on actions that will effectively and appropriately change the state to reach a goal, track an external process or reject a perturbation on the basis of a user-defined optimization criterion (often called objective or cost function). In fact, the definition of `controllability' is to be able to arbitrarily move a system from any state to any other state in finite time. Its counterpart is  `observability,' which is the formal quantification of the ability to arbitrarily extract the state of the system from measurements obtained by sensors \cite{kalman1959general,kalman1963mathematical}. When designing a robotic system, its state variables are explicitly chosen by the user from the many potential options, such as joint angles, angular velocities, endpoint locations, velocities, etc. Implementing observability via methods of state estimation is a form of system identification that spans a wide set of techniques, and which is central to any \fvc{control-theoretic} approach applied to robotics and models of biological function \cite{simon2006optimal,bryson1975applied,valero-cuevas2009computational}.

The field of neuromuscular control has come to the realization that it is still an open question whether and how the nervous system adopts the concept of state, and even whether it optimizes an explicit or implicit cost function \cite{deRugy2012muscle,loeb2012optimal,valero2015fundamentals}. This realization has been slow and reluctant because abandoning the sound engineering formalism is both unappealing and unnecessarily extreme. Thankfully, the past several decades have seen, as described below, the development of  equally well-founded alternatives or complements to the classical control theoretic perspective that mitigate the drawbacks of  state estimation, cost functions and even control laws. Furthermore, there are many other techniques and approaches that have yet to be applied to biological problems \cite{valero-cuevas2009computational}.

\subsection*{Feasible rather than optimal function}				
Optimization is a computationally efficient means to use convex, preferably quadratic, cost functions to select a specific control action from among all feasible actions within a high-dimensional solution space. For example, if you have N muscles crossing a joint, optimization can be used to search an N-dimensional space to find a point in it (i.e., a combination of muscle activations) to produce a given net joint torque while minimizing, say, the sum of squares of activations.

\fvc{Roboticists have always used optimization to control robots. Be it to tune gains in the P, PD or PID controllers  to implement force, impedance and position control \cite{ogata1997}, plan paths (e.g., \cite{shiller1989robot}), etc. In fact, the ubiquitous use of state estimators (such as Kalman Filters) are, in fact, optimal solutions in the least-square sense \cite{strang2003introduction}. In addition, this emphasis on optimality is central to \emph{Optimal Control} \cite{bryson1975applied} in its various modern forms as LQR, LQG, iLQR, iLQG (e.g., \cite{aschepkov2016optimal}.). From its inception, however, the emphasis on optimality has proven problematic to stability margins, which has led to other forms of control such as robust control, path integral control, model predictive control, etc. \cite{doyle1978,safonov2012origins,morari1993model, TheodorouJMLR}.}

Biomechanists and neuroscientists have, \fvc{ nevertheless, adopted the well-founded mathematical concept of optima} to cast the problem of neuromuscular  control as one of numerical optimization \cite{todorov2002optimal,Chao1978Graphical,crowninshield1981physiologically}. However, it is unlikely  that  the nervous system acts strictly like a computer running optimal control, gradient descent or policy gradient algorithms. Rather, optimization has and should be used as a metaphor, but one that should not be taken too literally when working with biological systems \cite{deRugy2012muscle,loeb2012optimal}.

We must recognize that, as with any metaphor, there are limits to its validity. It is important to explicitly acknowledge a bifurcation in the approaches we use to build robots vs. understand biological systems. For example, recent advances in control have begun to yield very impressive real-time performance in physical robots  \cite{williams2016aggressive,righetti2014autonomous,cifuentes2016probabilistic,kumar2015mujoco}---but there is no need to insist that biological systems use those methodologies or algorithms.

Nevertheless, scientists studying biological systems must ask themselves how faithful they want to adhere to physiological realism and, on the basis of that decision, select appropriate problems and solution methods---while also avoiding the temptation to necessarily imbue biological systems with mathematical algorithms, or robotic solutions with physiological meaning.
 
\fvc{An approach we can call \emph{Feasibility Theory} is an alternative to optimization. It tackles and solves the computationally expensive problem of explicitly defining and finding families of valid solutions (i.e., a `feasible muscle activation set')\cite{valero2015fundamentals}.} In doing so, we can characterize the set of options open to the nervous system without having to advocate a particular (and debatable) cost function \cite{valero2015fundamentals,valero1998large,valero2015exploring}. These families of feasible muscle activations have a well-defined, low-dimensional structure because they emerge, unavoidably and naturally, from the interactions among the known mechanical and physiological properties of the limb, and the functional constraints of the task (which can be mechanical, metabolic, physiological, etc.). That is, if you have a limb with 9 muscles and you are producing a task with five constraints (e.g., the x, y, z magnitude of the force vector produced by the endpoint of the limb, and the stiffness of the endpoint in the x and y directions), then the solution space is a 4-dimensional (i.e., 4=9-5) subset of the 9-dimensional  activation space \cite{kutch2012challenges,inouye2016muscle}.

More generally, thinking of the problem of controlling muscles as one of exploring and exploiting solution spaces is perhaps biologically tenable. Biological systems could use sparse trial-and-error learning to find and explore feasible activation sets. That is, implementing any point within the low-dimensional feasible solution space will be adequate to perform the task. Memory and pattern matching could be used to exploit the `information' collected from prior experience in the context of the current environment and task goals \cite{valero2015fundamentals}. 

\subsection*{Probabilistic sensorimotor control}
What information does the nervous system use to produce physical behavior? And how does it assemble, encode, store, access and use that information? A promising approach is a probabilistic one, where trial-and-error can be combined with memory to form probabilistic representations of actions in the physical world.  Bayesian inference  \cite{kording2006bayesian,peters2016size} or stochastic control \cite{todorov2002optimal} are formal approaches to describe the emergence of probability density functions of the mapping from perception/intention to action in the presence of unavoidable uncertainty, noise, risk and variability in the real world. Moreover, there are probabilistic approaches that can be used with populations of spiking neurons to produce physical behavior in a cost-agnostic, emergent, model-free way \cite{Sanger2016,dunning2015tuning,niu2017neuromorphic,jalaleddini2017neuromorphic}.

\fvc{Probabilistic approaches to control have their origins in robotics, where noise, uncertainty about the state, properties of the plant, dynamics of the coupled robot-world system, etc. have always posed challenges.} One can argue that such probabilistic models can be considered model-free or semi-model free because there is no explicit representation of the body or world, other than from the statistical representation of the results of trial-and-error experiments that inform the learning of a successful policy (i.e., control law or motor habit) that is valid in the neighborhood of  some initial conditions to meet a specific goal \cite{TheodorouJMLR,theodorou2011neuromuscular}.

Along these same lines, it is also possible that the controller itself is `embedded' in the structure of the hand. Thus, there  would be no need to regulate the manipulation task, but rather that the mechanical architecture of the hand naturally leads to adequate and versatile grasping function. This is sometimes called embedded logic or underactuated control, as described below \cite{Valero2007Anatomical,Rieffel2009Morphological,brock2016transferring,deimel2016novel,catalano2014adaptive,odhner2014compliant}.

\section*{Paradoxes and insights}
\subsection*{Under-determined vs. Over-determined mechanics}
One of the central tenets of of motor control has been the concept that the control of biological systems is under-determined, meaning that they have `too many' kinematic or muscular degrees of freedom. Therefore, the nervous system faces a problem of selecting and implementing a solution from among an infinite set of choices.

Such kinematic redundancy can be demonstrated by simple examples such as the possibility of using any one of multiple arm trajectories to hammer the same location in space \cite{Bernstein1967},  or one of multiple types of grasps to hold the object just as well \cite{Miller2004}. From the muscle control perspective, vertebrates have multiple tendons crossing each joint, thus, there are multiple individual muscle forces that can produce a given net joint torque \cite{Chao1978Graphical}.  In contrast, roboticists have emphasized design architectures to reduce kinematic and actuation redundancy and typically build robots with as few kinematic degrees of freedom or tendons to be controllable. 

This begs the question why the evolutionary process has tended to converged on such so-called under-determined mechanical systems for vertebrates. As reviewed in \cite{valero2015fundamentals}, thinking of biological systems as under-determined is paradoxical with respect to the evolutionary process and clinical reality. For example, why would organisms evolve, encode, grow, maintain, repair, and control unnecessarily many muscles when a simpler musculoskeletal system would suffice, and thus, have phenotypical and metabolic advantages? Why do people seek clinical treatment for measurable dysfunction even after injury to a few muscles, or mild neuropathology? Which muscle would you donate to improve your neural control?

Somehow, however, many muscles are a good thing. Given the evolutionary process, we probably have close to the right number of muscles to allow us to produce useful behavior in the real world \footnote{Note that this is not an argument for optimality of anatomical architecture, but only for sufficiency.}. One approach to  explain the apparent paradox that we have `too many' muscles in vertebrates is that every muscle expands our abilities and provides an additional degree of freedom for control. Behavior in the real world \footnote{Also called \emph{neuroethology} to distinguish it from reductionist laboratory work \cite{giszter2007primitives,ewart1980neuroethology}.} consists of satisfying multiple---at times competing---demands. Therefore, a mathematical argument can be made \cite{valero2015fundamentals} that behavior in the real world, by virtue of needing to satisfy multiple  demands or constraints, requires multiple muscles \cite{inouye2016muscle}. And, by extension, that dysfunction of even a few muscles will make the limb less versatile \cite{kutch2011Muscle}.

Similarly, kinematic redundancy loses its relevance when we consider that limbs are actuated by muscles that pull on tendons. It is clear that, if multiple tendons cross each joint, then there is redundancy in the sense that multiple individual muscle forces at those tendons that can produce a given joint torque. However, the same is not true when we consider movement. The rotation of that single joint defines the lengths of all muscles that cross it \cite{valero2015exploring,valero2015fundamentals,hagen2017similar}. While in principle muscles can go slack, muscles with tone will shorten appropriately. However, muscles that lengthen must lengthen by a prescribed amount. Thus, the relationship where a few joint angles and angular velocities for a given limb movement determine the lengths and velocities of all muscles is over-determined---the very opposite of redundant. That is, if \emph{any one} muscle that needs to lengthen to accommodate the movement fails to do so (because, for example, it received the incorrect neural command or its stretch reflex fails to be appropriately modulated in time), the movement will be disrupted \cite{valero2015exploring,valero2015fundamentals,hagen2017similar}. Therefore, while multiple limb kinematics may be equivalent from a task perspective (i.e., reaching a cup or throwing a ball), they are far from equivalent from the perspective of neurophysiological control, robustness to sensorimotor noise, or time-critical modulation of activation and reflexes \cite{hagen2017similar}.

\subsection*{Grasp vs. Manipulation}
Another recent evolution in the biological side has been the explicit distinction between grasp and manipulation. Although these terms are often used interchangeably in the biological literature, there is a long tradition of creating clear taxonomies and descriptions of hand actions that clearly distinguish between the two \cite{cutkosky1989grasp,mackenzie1994grasping,brand1993clinical,murray1994mathematical}. Specifically, grasp in general relates to the act of seizing an object by wrapping the fingers around it. Manipulation has a more general connotation of imparting change to an object or process. However, precision or dexterous manipulation is a more specific term reserved for cases where only the fingertips make contact with the object, not simply to grip the object, but to be able to act independently to produce in-hand manipulation.

Interestingly, most biological research has focused on grasp \fvc{\cite{brock2016transferring,murray1994mathematical,cutkosky1989grasp}}. Largely because studying precision or dexterous manipulation has important practical difficulties with motion capture of individual finger motions, and measurement of individual fingertip forces. Similarly, in spite of the mathematics of dexterous manipulation being well developed \cite{murray1994mathematical}, robotic hands and prostheses tend to focus on grasp because of the difficulties in designing, building, and controlling fingers and finger contacts independently.

There have been some important advances, however. For example, it is possible to begin to simulate contact forces that go beyond physics engines for gaming or animation \cite{kumar2015mujoco}. Similarly, some experimental methods have been developed to quantify dynamic dexterous manipulation, which has revealed novel aspects about the neuromechanical control of dexterous manipulation in development, adulthood, healthy aging and neurological conditions (for overviews see \cite{Valero-Cuevas2003strength-dexterity,lawrence2014quantification,ko2015force,lawrence2015outcome,pavlova2015activity,mosier2011controlling}), as well as how the interaction between cognitive and biomechanical factors affects dexterous manipulation performance (e.g., \cite{Fu2010,Fu2011,Fu2015}).  

\subsection*{Under- vs. Over-actuated control}
Similarly, it is reasonable to ask to what extent the nervous system is necessary for grasp (and perhaps even manipulation). It is an analogous question to what has also been recognized for passive dynamic walkers \cite{mcgeer1990passive,collins2005efficient}. While effective mechanical function can be found in many vertebrates, primates (and humans) are the beneficiaries of highly specialized neuroanatomical coevolution of brain and hand (e.g., \cite{lemon2008descending,Schieber2004,schieber2001constraints,wilson2010hand}). Understanding the contributions of a neural controller, or specific neuroanatomical areas of the brain, to grasp and manipulate remains an active area of study. In fact, it is critical to consider moving away from a strictly somatotopic \cite{sanes2001orderly} and cortico-centric view of manipulation, especially in the cases of dynamic dexterous manipulation where time delays preclude active involvement \cite{lawrence2014quantification}.

After all, the current concept of cortical control is not the exclusive micromanagement of individual muscle activations, but rather includes the `binding' of motor neurons into flexible, context-dependent functional groups \cite{nazarpour_flexible_2012,farmer_rhythmicity_1998,de_vries_functional_2016}, the utilization of primitive `synergies' prepared by networks of spinal interneurons \cite{rathelot_subdivisions_2009}, adjustment of sensory feedback gains \cite{shadmehr_computational_2008}, and the formation/recall of motor memories \cite{nozaki_tagging_2016}, to name just a few.  Synergies are discussed in more detail in subsequent sections.

Nevertheless, as in the case of passive dynamic walkers, robotics provides counterexamples to such micromanagement of muscle actions by the brain, or even the nervous system in general. A class of robotic hand designs is called under-actuated because few motors drive multiple degrees of freedom (this is in contrast with over-actuated hands that have enough muscles to control every degree of freedom independently). Such hands can display multiple versatile grasp functions, without requiring a controller \cite{odhner2014compliant, deimel2016novel} or even fingers \cite{brown2010universal}.  Such developments are alternatives that promise to develop multiple designs along the spectrum between under- and over-actuated robotic hands. This is especially useful in cases of brain-machine interfaces for hand prostheses, where only a few degrees of freedom of control can be extracted from the human pilot's nervous system (e.g., \cite{aflalo2015decoding}). 

\subsection*{Learning vs. Implementation vs. Adaptation} 
Human sensorimotor learning has been extensively studied \cite{Wolpert2011,wolpert2001perspectives}. One view posits that that humans' ability to perform skilled motor behaviors relies on learning both control and prediction through inverse and forward internal models (implicit, explicit, probabilistic or otherwise). Specifically, a given control strategy generates motor commands needed to create desired consequences (e.g., a given reach trajectory or grasping an object at specific locations), whereas prediction maps motor commands (i.e., efference copy) into expected sensory consequences (e.g., object contact or onset of acceleration at object lift off \cite{Flanagan2003,Haruno2001}). The mechanisms proposed to account for updating of these internal models may or may not include errors that would occur when a mismatch between sensed and predicted sensory outcome occur, i.e., error-based learning (e.g., \cite{Kording2007,Krakauer2005,Thoroughman2000}) and use-dependent learning (e.g., \cite{Classen1998,Diedrichsen2010,Verstynen2011,Huang2011}). However, most of what we know about human sensorimotor learning for reach has been derived \fvc{from studies of reaching movements over distances of $\pm$ 10---14 cm, and  their adaptations} to force fields or visuomotor rotations. Relatively little is known about mechanisms underlying sensorimotor learning of grasping and manipulation.

We have known for decades that finger force control used in previous manipulation can influence how forces are coordinated on the current manipulation through the memory of an object's physical properties \cite{johansson1984roles,Johansson1988,Johansson1992}. More recently, it has been shown that humans may acquire and retain multiple internal representations of manipulation \cite{Quaney2003,Quaney2005}. Later studies provided further evidence supporting the concept of multiple sensorimotor mechanisms and how their different time scales may interfere with generalization or retrieval of previously learned manipulation \cite{ingram2010multiple}---even when the object being manipulated is the same. A recent study has provided evidence for the co-existence of context-dependent and independent learning processes \cite{Fu2015}, which would operate similarly to those described for adaptation of reaching movements \cite{Lee2009Adaptation}. The advantage of context-dependent representations of manipulation is that they can be recalled when the object has strong contextual cues (i.e., object geometry and perhaps other perceptual attributes). In contrast, context-independent representations are more sensitive to the practice schedule used to learn a given manipulation, but might be particularly advantageous when the upcoming context has no context cues. That is, in the absence of information to the contrary, it is preferable to repeat the most recent manipulation strategy even though it is not guaranteed to be the correct one.

When considering parallels between the above-described framework for learning of dexterous manipulation in humans with learning manipulation by robotics systems, it has been suggested that artificial controllers could take advantage of select features of the biological framework. Specifically, and as reviewed in \cite{Fubookchapter}, multiple parallel learning mechanisms could benefit robotic learning of manipulation tasks to afford to deal with structured and unstructured environments. At the same time, the detrimental effects or interference of neural representation built through learning \fvc{in} one manipulation context, \fvc{and then transferring it to} another context can be theoretically minimized or bypassed when designing an artificial controller. \fvc{Some examples of successful robotic learning for grasp and manipulation show that this is possible \cite{cifuentes2016probabilistic,righetti2014autonomous,kemp2007challenges,saxena2007robotic}.} Of course, these theoretical considerations assume that building multiple representations of learned manipulations allows them to operate independently, something that---as described above---clearly \fvc{also} challenges biological controllers. 

Another biologically-inspired phenomenon that could be of value to robotic manipulators is finger force-to-position modulation. Briefly, it has been shown that humans are able to modulate manipulative forces in an anticipatory fashion, i.e., between contact and onset of manipulation, according to where the object is grasped \cite{Fu2010,Fu2011}. This phenomenon, which has been confirmed by several studies \cite{Fu2014,Fu2014a,Mojtahedi2015,Marneweck2016}, ensures attainment of the manipulation goal despite trial-to-trial variability in finger placement that may naturally occur while using the same or different number of fingers (\cite{Fu2010} and \cite{Fu2011}, respectively). Finger force-to-position modulation is a phenomenon that is very useful for inferring its underlying neural control mechanisms. Specifically, for humans to be able to adjust finger forces as a function of variable position, a `high-level' representation of the task (e.g., a given compensatory torque) is required, rather than learning a fixed finger force distribution. Additionally, such high-level representation has to drive how sensing of the relative position of the fingers is used to implement the appropriate finger force distribution by the time the learned manipulation is initiated. As finger force-to-position modulation affords biological systems to be very adaptive---a given manipulation can be performed without having to grasp an object exactly in the same way each time it is being manipulated---one can envision important robotics applications. These include controllers that are designed to build, through extensive training, the high-level representation of a task performed in many different ways. If such high-level representation could be built, stored, retrieved, and designed to interact with artificial sensing of finger positions, such a controller should, theoretically, be able to be adaptable to manipulators that widely differ in terms of number of joints or fingers. Such a controller could be shared by multiple representations learned through training of manipulation in structured and unstructured contexts.
 
Another important distinction to be made is that, as roboticists, we marvel at the learned capabilities of biological systems. However, we tend to forget how difficult it is for organisms to learn and maintain that level of performance. Recent work has begun to elucidate why learning to produce accurate, smooth and repeatable movements takes immense amounts of practice even in typically developing children \cite{adolph2012you}, why so few of us can become elite musicians or athletes \cite{gladwell2008outliers}, and why rehabilitation requires very intensive practice \cite{lohse2014more}.  That is, controlling our bodies is not as easy as it appears. We are seeing the result of millennia of co-evolution and years of development, training and learning. Moreover, in the case of manipulation, we have co-evolved environments, objects and tools to match the capabilities of our hands. The design of airplane cockpits, left- and right-handed scissors, frets in string instruments, the key system in clarinets, and touch screens are but a few examples.

Thus, biological hands in particular have an unfair advantage over robotic hands and prosthetics. Engineers should explicitly begin to decide what functionality and control to embed in the mechanics of the system, what control algorithms to use for learning vs. standard performance vs. elite performance vs. adaptation. It is not unreasonable to propose that robotic hands, once built, should undergo a developmental learning process (a `robot kindergarten?') to learn the specific control algorithms, motor habits, and statistically useful anticipatory strategies defined by their intended use or---in the case of prosthetics---the environment, job and preferences of their human pilot. Insisting on a one-size-fits-all, real-time control approach to robotic hands has been shown to be overly ambitious, and even unnecessary \fvc{as demonstrated by the capabilities of the under-actuated hands mentioned above (e.g., \cite{catalano2014adaptive,deimel2016novel}), as well as grippers with no fingers at all \cite{brown2010universal}.} \fvc{Some salient examples of such learning (and re-learning) come from \cite{bongard2006resilient,kalakrishnan2011learning,bristow2006survey}.}

\subsection*{Prescriptive vs. Descriptive synergies}

\fvc{What are the debates in the study of synergies in biological systems? A root cause of the debates is the nature of scientific inference based on experimental observations. The fact that experimental recordings detect dimensionality reduction is not surprising because sensorimotor control must, by definition, select motor actions from within the low-dimensional subspace of feasible actions \cite{kutch2012challenges}. Therefore, disambiguating \emph{prescriptive} synergies of neural origin (those that are prescribed by the nervous system as a control strategy) from \emph{descriptive} synergies (those that describe the expected dimensionality reduction) is difficult \cite{kutch2012challenges,brock2016transferring}. Thus, the main question is not \emph{whether} the nervous system inhabits a low-dimensional solution space to perform tasks, but rather \emph{how} it does so \cite{Valero-Cuevas2009Structured,racz2013spatio,kutch2012challenges,tresch2009Case}. Moreover, although several tasks can share the same general features captured by such dimensionality reduction, it is perhaps the fine details particular to each task that may be critical to their performance \cite{brock2016transferring}.}

Biological `controllers' co-evolved with mechanical systems whose operations are characterized by a very large number of elements---e.g., motor units, muscles, and joints---while relying on their spatiotemporal coordination and adaptability to task demands. The refinements afforded by such evolution can be appreciated when examining the efficacy of the neural control of several complex motor behaviors, including, but not limited to, speech production, locomotion, and manipulation. When examined in detail, researchers were surprised and intrigued to see that those functionally complex behaviors that involve the control of many variables (like the 20+ angles for the joints of all fingers) in reality evolve in a lower-dimensional space (i.e., can be well approximated by roughly 5 variables) \cite{santello1998postural}. Such motor \emph{synergies} have also been observed in the phase-locked coordination (or correlated action) of multiple muscles that produce complex behaviors  \cite{giszter1989kinematic,scholz1999uncontrolled,giszter2007primitives}

The theoretical framework of synergies has been extensively used and tested to account for the nervous system's ability to control multiple muscles and multi-joint movements (for reviews see \cite{Ting2007,Bizzi2008,Lacquaniti2013}). Synergies would operate by constraining the spatial and temporal activation of multiple muscles. Therefore, the existence of consistent covariation patterns in electromyographic (EMG) activity or joint excursions, whose structure can be spatially and/or temporally modulated according to task requirements, would be compatible with the synergy framework. Synergies have also been used as a framework to understand pathological coordination of movement (for reviews see \cite{Cheung2012,santello2015movement}). However, longstanding issues remain regarding the extent to which synergies can be considered `fixed' building blocks of movements, the extent to which they are modifiable as a function of task demands, adaptation and perceptual context, as well as their very role in facilitating sensorimotor learning in tasks that may benefit from, or be penalized by a synergy-like control structure \cite{mechsner2001perceptual} ({for a review see \cite{Santello2013}).

\fvc{When considering the biomechanics of hand muscles, the existence of anatomical constraints would support synergistic actions of the fingers. These constraints can come from, for example, finger muscle-tendon complexes spanning several joints and passive linkages among tendons \cite{schieber2001constraints}.} Such synergistic actions have been described as subject-independent finger kinematic patterns for grasping \cite{santello1998postural,Santello2002} (for review see \cite{Santello2013}), as well as coupling of finger movement or forces among non-instructed fingers when humans are asked to move or exert force with one finger \cite{Reilly2000,Zatsiorsky2000} (for review see \cite{Schieber2004}). Early attempts to define the control of individuated finger forces in cortical neuron activity revealed a much more complex picture characterized by broadly distributed activity \cite{schieber1993somatotopic}. More recent work in non-human primates, however, supports an organization of cortical activity that is compatible with the synergy framework \cite{Overduin2012}. When searching for neural correlates of synergies in humans, a recent study revealed that the cortical representation of hand postures can be better accounted for by using a synergy-based network than somatotopic or muscle-based models \cite{Leo2016}, which is compatible with the view of cortical organization of finger movement being shaped by habitual use \cite{Ejaz2015}, and even goal equivalence in finger actions being implemented at a cortical level \cite{babikian2017cortical}.

Many studies have attempted to identify synergies at different levels of biological systems and species, including primary motor cortex \cite{Leo2016}, spinal cord \cite{Giszter2013}, motor units \cite{Winges2004,Santello2004,Winges2008}, motion \cite{flanders1992kinematics,santello1998postural}, and forces \cite{Santello2000,Rearick2003,zatsiorsky2003prehension}. However, the functional role of synergies has been debated for decades, partly due to the fact that the operational definition of synergies can vary significantly depending on several factors, including the level of the system at which they are analyzed, the methods used to quantify them, and the tasks used to prove or negate their existence (as discussed in \cite{Santello2013}; see also \cite{Santello2016}). Among the conceptual frameworks that have been proposed, synergies would be instrumental in reducing the number of independent degrees of freedom that the nervous system has to control as originally proposed by Bernstein \cite{Bernstein1967}, or ensure attainment of task goals by minimizing the variance that would be detrimental to performance \cite{scholz1999uncontrolled,scholz2003uncontrolled,Kang2004}. An alternative interpretation of the role of synergies, however, points out the difficulty of interpreting synergies as the root cause of multi-muscle coordination or a byproduct of mechanical interactions between the biological system and the environment \cite{tresch2009Case}.

Robotics, in contrast, synthetically designs, assembles and operates engineered systems where synergies can be prescribed. Over the past two decades, roboticists have exploited the concept of (prescriptive) synergies to design robotic hands (for review see \cite{Santello2016}). Examples of these designs include the Pisa/IIT SoftHand \cite{catalano2014adaptive}, whose design was based on the kinematic synergies extracted from grasping a set of imagined objects \cite{santello1998postural}, as well as devices to constrain motion of human fingers for rehabilitation of sensorimotor function \cite{xiloyannis2016modelling}. Here, the underlying design motivation is to capture human-like kinematic features, i.e, simultaneous motion of all fingers, by using a significantly smaller number of actuators than joints. Preliminary clinical applications of this approach for prosthetic applications have shown that individuals with upper limb loss can quickly adopt such synergy-based design with minimal training \cite{zhao2015application}. A major goal and challenge for robotic grasping and manipulation is the implementation of force control using kinematic synergies. The results of computational modeling suggest that the first few hand postural synergies may play an important role for attaining force closure \cite{bicchi2011modelling}. Nevertheless, it remains to be investigated the extent to which robotic motion-to-force transition can fully leverage a synergy-based motion-to-force coordination. Experimental evidence and theoretical frameworks developed by studies of human multi-finger synergies might potentially be used to inspire a hierarchical control of high- and low-level grasp variables (i.e., task goal and distribution of individual fingertip forces, respectively) \cite{Fu2011}, as well as `default' vs. task-dependent modulation of fingertip force distributions \cite{Rearick2003}. Nevertheless, a major challenge, both from neuroscientific and robotics perspective, is evaluating the role of sensory feedback elicited by object contact and force production on the coordination of multiple (human and robotic) fingers.

\section*{Conclusions}

The literature on the biological and robotic approaches to grasp and manipulation is large, and has experienced exponential growth in the past two decades. The reinvigorated interest in this topic has come, as in past conflicts, from governments  attending to wounded soldiers and civilians who survive traumatic loss of limbs (e.g., the DARPA program to Revolutionize Prosthetics), from the need to improve the quality of life of individuals with stroke, cerebral palsy and other neurological conditions that now have greater survival rates, and from recent advances in autonomous and humanoid robots for whom manipulation remains a litmus test for performance.  A positive feature of the latest developments has been the greater and fruitful cross-fertilization between biology and robotics approaches. \fvc{It is no longer an aberration to have engineers who are well versed in neuroscience working in robotics or neuroscience, or neuroscientists who are well versed in engineering working on scientific problems or robotics.} This greater interaction across fields, however, has the added burden of needing to understand and keep current on two vast and rapidly growing fields---which has led to confusion of terms and principles, duplication of efforts, loss of nuance in translation, and lack of familiarity with fundamental concepts.

Thus, our goal here is to provide researchers working in these fields the briefest of overviews of advances (conceptual and material), pitfalls, and open questions. 
\fvc{In particular, we aimed to provide specific examples to clarify when and why a better understanding of the biology of grasp and manipulation would benefit robotic grasping and manipulation. Namely, the evolutionary process has yielded organisms that can produce versatile function even when cherished engineering principles are not present and biological systems operate well in spite of having noise, delays, nonlinearities, etc. There are many lessons learned, described above, such as hierarchical and open-loop control, morphology that simplifies the control problem, the utility of having multiple muscles, etc.  Conversely, engineering thinking has provided well-founded principles of mechanics, mathematics and control engineering to aid the scientific work aimed at understanding the abilities of human hand.} Some salient points are:
 
\begin{itemize}

\item Deductive vs. synthetic science\\
Engineering approaches to understand biology---and biology as an inspiration to engineering---have generated a rich repertoire of experimental and theoretical advances in both areas. However, it is important to be aware of fundamental differences and inherent limitations \fvc{of scientific} vs. synthetic approaches. It is the recognition of these differences that needs to guide work in each field and in their interactions. This will allow those interactions to be as fruitful as possible.
 
 \item Redundancy\\
The mathematics of robotics has contributed greatly to our ability to study biological limbs and hands. Unfortunately, the joint-torque formulation---while valid and correct---can oversimplify the nature of the problem that confronts the nervous system. At its core, the control problem the nervous system faces is one of linear actuators (muscles) that can only actively pull on tendons. We have revisited the redundancy problem while also emphasizing that the control of forces is distinct from the control of movements.
This has allowed us to clarify many aspects of biological function and dysfunction.  For example, it is now clear that muscle redundancy to produce a set of net joint torques is not as great as once thought. This is because the set of feasible muscle activations has a very strong structure. Similarly, kinematic redundancy is greatly constrained because different movements are not entirely equivalent. From the perspective of muscle excursions and the regulation of stretch reflexes, each natural movement in fact represents an over-determined problem (the opposite of redundant), where muscle activations and reflex modulation must follow very specific spatiotemporal patterns.
This all begins to explain why it is so difficult to learn to use our limbs and hands (i.e., it takes years of practice), why dysfunction arises even from minor neuromuscular damage, why rehabilitation is so difficult, and why limbs have evolved to have `so many' muscles. Roboticists can use these lessons to revisit their design specifications, which often prefer torque motors or a sparse architecture where each joint is controlled by two dedicated tendons. 

\item From motor towards sensorimotor\\
Sensory feedback plays critical roles in human grasp and manipulation for building internal representations that are considered critical for predicting sensory consequences of hand-object interactions; and for online sensing of those interactions. This dual nature of sensory feedback (e.g., exploration-exploitation; learning-execution) is particularly challenging to capture in robotics. From the technical perspective, replicating human tactile function has proven difficult, although there have been recent advances not only in sensor hardware development; but also in the expansion of prostheses to explicitly add hand proprioception and touch interfaces (i.e., DARPA's HAPTIX program). But equally important, the field has begun to recognize that the use and interpreting sensory function may change as the learning process evolves. Thus, as in biological systems, on-line monitoring and processing of sensory information is necessary at first, but can actually be counterproductive and slow down the performance once learning has stabilized. Thus, hierarchical, distributed and state-dependent gating, processing and use of sensory information is yet to be well understood in human hands, and should be a goal of robotic systems.

\item Grasp is not manipulation\\
Grasp and dexterous manipulation, although often interchangeably referred to in neuroscience literature, are fundamentally different and challenge biological and robotic controllers in unique ways. In robotics, greater advances have been made in grasping than manipulation. Some important advances are being made to understand and implement the ability of fingertips to produce motion-to-force transitions, force-to-position modulation, and in-hand object re-orientation/re-configuration, mostly in simulation \cite{kumar2014real,fu2013human}. The challenge is now how to design and build robotic hands with the mechanical, motor and sensory abilities to implement those behaviors. This is a very promising and necessary direction for research. On the biological front, the new approaches to understand dynamic dexterous manipulation with the fingertips allows the quantification of performance. But more importantly, those findings challenge the dominant cortico-centric view of manipulation, and highlight the importance of considering subcortical, spinal and neuromechanical contributors. This further motivates the exploration of hierarchical and distributed mechanisms that allow such versatile capabilities in the presence of delays, noise and uncertainty.  Such evolution in our thinking promises to revolutionize our understanding of the mechanisms that enable healthy function, explain disability, and  inform rehabilitation. Combining future developments along these parallel lines of work also promises important developments in the evolution of the design and control of robotic and prosthetic hands.

\item Neuromorphic vs. neuromimetic\\
Unlike neuromimetic approaches aimed at capturing aspects of neural function without adhering to the replication of biological structures or processes, neuromorphic approaches seek to replicate (to the extent possible possible) the biological components and mechanisms at a particular scale. This allows us to ask whether and how neuromechanical function can emerge without being defined \emph{a priori}. This allows us to understand the extent to which the implementation of the structures and processes (i.e., spiking neurons, delayed transmissions, muscle nonlinearities) define and contribute to function, and the different presentations of dysfunction. Thus, neomorphism is proposed as a means to reconcile engineering and biology, as well as accelerate their cross-fertilization.

\item Prescriptive vs. descriptive synergies\\
Synergies have been defined and studied by neuroscientists as building blocks of complex movements based on observations made at different levels of the neuromuscular system. Although their fundamental characteristics and functional role remain to be established, the concept of synergies has inspired robotic and prosthetic hand design not only to simplify control, but also to capture human-like motion features---the latter objective being particularly important for the assistive technologies where there are few degrees of freedom for control and the prosthetics are underactuated. From the conceptual perspective, however, it is important that we recognize that the execution of a task will always exhibit dimensionality reduction in the control, kinetic and kinematic variables because they are, by definition, inhabiting the subspace of feasible actions that satisfy the constraints that define the task. Therefore, it is to be expected that (descriptive) synergies will be detected when studying human function. Nevertheless, interesting, valid and open questions remain in our understanding of whether and how the healthy and damaged nervous system implements (prescriptive) synergies to inhabit those subspaces (i.e., the feasible activation manifold for a given task).

\end{itemize}

It is our hope and expectation that these recent conceptual clarifications, advances and newly-defined open problems will accelerate our understanding of healthy and pathologic hand function (and biological function in general), and will catalyze the creation of truly versatile and dexterous robotic hands and prostheses.

\section*{Declarations}
\subsection*{Ethics approval and consent to participate}
Not applicable in this review article.

\subsection*{Consent for publication}
The authors consent.

\subsection*{Availability of data and material}
Not applicable in this review article.

\subsection*{Competing interests}
FV-C holds US Patent No. 6,537,075 on some of the technology reviewed in this article that is commercialized by Neuromuscular Dynamics, LLC. MS has no financial or personal relationships with other people or organizations that could inappropriately influence this work.

\subsection*{Funding}
Research reported in this publication was supported by the National Institute of Arthritis and Musculoskeletal and Skin Diseases of the National Institutes of Health (NIH) under award numbers AR050520 and AR052345 to FV-C, and Eunice Kennedy Shriver National Institute Of Child Health and Human Development of the NIH under award number HD081938, National Science Foundation (NSF) under award number BCS-1455866, and the Grainger Foundation to MS. The content is solely the responsibility of the authors and does not necessarily represent the official views of the NIH or NSF.

\subsection*{Authors' contributions}
FV-C and MS contributed to the conceptualization and writing of this review article.

\subsection*{Acknowledgements}
We thank Dr. Qiushi Fu, Dr. Christopher Laine, Dr. Kian Jalaleddini, Suraj Chakravarthi Raja, Ali Marjaninejad,  Melisa Osborne, and Jamie Flores for their help with critical reading of this manuscript.
\newpage
\section*{References cited}




\bibliographystyle{model1-num-names}
\bibliography{BANCOM6.bib}







\end{document}